The Hangulphabet: A Descriptive Alphabet

Robert Bishop
and
Ruggero Micheletto

Yokohama City University
{rbishop}@yokohama-cu.ac.jp
Yokohama City University
International College of Arts and Sciences
22-2 Seto, Kanazawa-ku,
Yokohama-shi, 236-0027 Japan

Abstract

This paper describes the Hangulphabet, a new writing system that should prove useful in a number of contexts.  Using the Hangulphabet, a user can instantly see voicing, manner and place of articulation of any phoneme found in human language. The Hangulphabet places consonant graphemes on a grid with the x-axis representing the place of articulation and the y-axis representing manner of articulation.  Each individual grapheme contains radicals from both axes where the points intersect.  The top radical represents manner of articulation where the bottom represents place of articulation.  A horizontal line running through the middle of the bottom radical represents voicing.  For vowels, place of articulation is located on a grid that represents the position of the tongue in the mouth.  This grid is similar to that of the IPA vowel chart (International Phonetic Association, 1999).  The difference with the Hangulphabet being the trapezoid representing the vocal apparatus is on a slight tilt.  Place of articulation for a vowel is represented by a breakout figure from the grid.  This system can be used as an alternative to the International Phonetic Alphabet (IPA) or as a complement to it. Beginning students of linguistics may find it particularly useful. A Hangulphabet font has been created to facilitate switching between the Hangulphabet and the IPA.

## Introduction

The Hangulphabet is a new alphabet that represents both place and manner of articulation in one grapheme.  As its name implies, it was inspired by Hangul; the system used for writing the Korean language.  Among other things, the Hangulphabet can be used as an alternative to the IPA.  I am confident it will greatly facilitate a number of tasks in the field of linguists, especially those new to the science.

## Origins

The idea for the Hangulphabet first occurred to me when I started studying linguistics. When finding answers to phonology and historical linguistics problems, I spent lots of time looking back and forth between the IPA symbols and what they represented on the official chart.  To me, symbols like ɕ and ɣ didn't tell me a lot about the pronunciation of their respective phonemes.  I am aware that for seasoned linguists, a grapheme such as "ɳ" signifies a nasal retroflex but for a neophyte, characters like these just added to the burden of getting a feel for how see phonological patterns.

**Hangul and the Hangulphabet**

In one of my early phonology courses, one of my professors mentioned that the Hangul writing system actually represents what happens during the articulation of different sounds.  For example: the glyph "ㅁ" represents the bilabial nasal /m/ the glyph "ㅂ" represents /p/, a bilabial stop. Both symbols incorporate a square which signifies the same place of articulation; in this case bilabial (Sek Yen Kim-Cho, 2002). This system works very well for the Korean language and makes learning to read much easier that the Chinese character system that preceded it.  One can just mimic the symbols with one's mouth (as I will describe further below).   The following quote best describes Hangul and its simplicity; "A wise man can acquaint himself with them (Hangul glyphs) before the morning is over; a stupid man can learn them in the space of ten days." (Haerye, 1449 pg. 27a).

Using Hangul as inspiration, I have created an alphabet that can represent exactly where and how a sound is articulated.  There are a few notable differences between my system and Hangul that I should mention. Unlike Hangul, the Hangulphabet is not written in moraic blocks.  Moraic blocks work well for Korean which is mora based and has few consonant clusters but not for other languages like English or Russian. Even if the Hangulphabet were to be used solely for mora based languages, the characters would be too large to be written in blocks.

Although the place of articulation is very clear in Hangul, manner of articulation is often implied.  For example, The grapheme "ㄴ" (/n/) is an alveolar nasal. Although the curved line represents the position of the tongue (if one is looking to the left), the fact that it is nasal is not noted.  Another issue that makes this system difficult to expand to other languages is the fact that there are many voiced and unvoiced allophones in Korean. For example, "ㄱ" is a velar plosive.  Once again, as the grapheme beautifully and simplistically shows the tongue is now against the velum in the back of the mouth. To Korean speakers however, this could be either a [k] or a [g] depending on the surrounding sounds. In most cases it wouldn't be distinguished.  In short, like any other language, Hangul works best when representing the language it is intended to represent.

**How the Hangulphabet is used**

In contrast, graphemes in the Hangulphabet represent both place and manner of articulation.  The top radical in a grapheme represents the manner of articulation.  The bottom radical represents the place of articulation. A line running through the middle of the bottom radical represents voicing. An example would be the grapheme "ㅁ̌" (/p/). The radical "×" on the top denotes the fact that the sound is plosive.  The square radical at the bottom of the grapheme shows that the sound is bilabial.  Like the grapheme "ㅁ̌" (/p/), the letter "ㅂ̌" (/b/) shares everything in common with the exception of voicing thus the only difference in these graphemes is the line running though the bottom radical. Furthermore, if we once again change the top radical to a nasal, "ᵛ" we get "ㅂ̌" (/m/).

Table 1 below is based on the consonant chart from the IPA (International Phonetic Association, 2005).  Using this chart, you can put both manner and place of articulation on a grid.  Using this grid, only 22 radicals need to be memorized as opposed to the 64 that currently exist on the IPA chart.   Just learning the 10 most high frequency glyphs such as those representing fricatives and alveolars can go a long way in helping a new linguist identify sounds and patterns in a short time.  Using this system, one could also use characters that are 'unofficial' by combining two radicals.  Although colleagues may disagree with the existence of a sound, they would know what the grapheme representing it meant.

    Table 1. The Hangulphabet Consonant Chart

| CONSONANTS (PULMONIC) | LABIAL | | CORONAL | | | | DORSAL | | | | RADICAL | | LARYNGEAL |
|---|---|---|---|---|---|---|---|---|---|---|---|---|---|
| | □ Bilabial | ⊔ Labio-dental | ▯ Dental | ⊏ Alveolar | ⊤ Palato-alveolar | ⌐ Retroflex | ⊤ Alveolar-palatal | ʃ Palatal | ⊐ Velar | ⊐ Uvular | ∧ Pharygeal | ∧ Epi-glottal | ⊼ Glottal |
| Nasal | m̄ m | m̄ ɱ | | n̄ n | | n̠̄ ɳ | | ɲ̄ ɲ | ŋ̄ ŋ | N̄ N | | | |
| Plosive | p̄ p, b | | | t̄ t, d | | t̠̄ ʈ, ɖ | | c | k̄ k, g | q̄ q, G | | ʔ | ʔ |
| Fricative | β, ɸ | f, v | θ, ð | s, z | ʃ, ʒ | ʂ, ʐ | ɕ, ʑ | ç, j | x, ɣ | χ, ʁ | ħ, ʕ | H, ʕ | h, ɦ |
| Approximate | | ʋ | | ɹ | | ɻ | | j | ɰ | | | | |
| Tap, flap | | ⱱ | | ɾ | | ɽ | | | | | | | |
| Trill | B | | | r | | | | | | | | | |
| Lateral fricative | | | | ɬ, ɮ | | | | | | | | | |
| Lateral approximate | | | | l | | ɭ | | ʎ | L | | | | |
| Lateral flap | | | | ɺ | | | | | | | | | |

    There are other features that simplify things even further. For the most part, the glyphs used to represent manner of articulation show the position of the tongue across the palate. Looking at the horizontal axis, moving from left to right, you'll see that after the first three glyphs, the perpendicular line likewise moves from left to right. This line represents the position of the tongue across the palate during articulation.  The far left starts with the front of the palate moving back as you move right. The first two glyphs involve squares as the tongue is not involved in the production of these sounds. The symbol for dentals (▯) acts as a transition between the labials and sounds produced with the tongue.  Also, A character that didn't include a perpendicular was preferable in this case due to spacing and readability issues.

    The vertical line of the grid represents manner of articulation.  Although the glyphs are abstractions, many can be easily identified.  The symbol (×̂) to me best represents plosives as it resembles a little explosion. For older linguists it also looks like a little signal to 'stop'. The nasal symbol (‿) resembles a nose.  The approximate symbol (⊓) portrays a person holding his hands to display the approximate size of an object.  The other symbols, unfortunately, will take a little more mnemonic creativity.

    Vowels are similar in many respects.  Each symbol represents place of articulation.

Below is the Hangulphabet vowel chart. Each glyph represents a section of the mouth where the sound is produced. In my version of the vowel chart, I have tilted the trapezoid commonly used to represent vowel articulation. I have made this adjustment to make the resulting glyphs more distinguishable. Thus the front unrounded vowel " i" (\), the mid unrounded vowel "ɨ" (|) and the back rounded (or unrounded for that matter) "u" (/) are readily seen. If you notice, the line coming out from the side represents rounding. If the line is to the left, the sound is unrounded, to the right, it's rounded.

Table 2 The Hangulphabet Vowel Chart

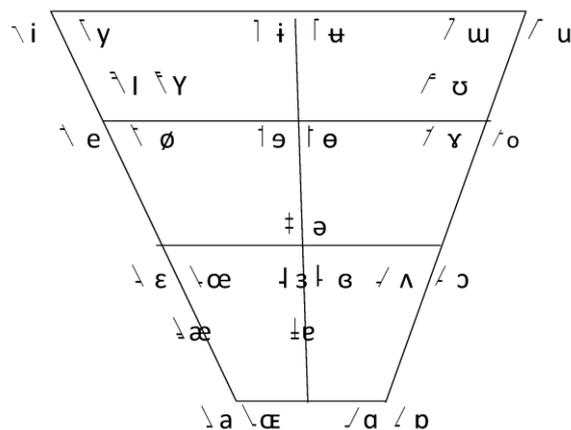

As for diacritics, the standard IPA diacritics can be applied to this system. The same can be said for sounds that lie outside the place and manner of articulation grid such as clicks. These graphemes work the same with the Hangulphabet.

**The Hangulphabet and Visible speech**

Before going further, I should mention that I am aware of the existence of Visible Speech, a system developed in 1864 by Alexander Melville Bell for the deaf. It very closely resembles the system I am proposing. I discovered this system long after I started work on the Hangulphabet. Had I discovered this earlier, I may have considered making some modifications to fit the needs of linguists rather than using Hangul or mixing the two. I believe however that although my system is not as aesthetically pleasing as Visible Speech, its letters are clearer and easier to discern.

**Hangulphabet and the IPA**

The International Phonetic Alphabet has always served the linguistic community well. Many of its symbols, however, are unfamiliar to the uninitiated. In addition to this issue, all of the symbols in The IPA to my knowledge come from Western writing systems. Although this came about as a result of accidents in history, The Hangulphabet by comparison, is not clearly associated with any geographical location. It is loosely connected with a system unique to Korean which has no history to my knowledge of linguistic imperialism. Such a system should find some favor for linguists around the world.

**The Hangulphabet Font**

Along with this chart and system, I have created a new font that will make its use more feasible.  This font will also allow one to easily toggle between the IPA and the Hangulphabet. Although it was created on a Macintosh, so far it only works with Microsoft Word on a Windows operating system.  Once downloaded and installed, all one needs to translate IPA to the Hangulphabet is to switch between fonts.  The font I made is a modification of SIL IPA Doulos taken from the SIL institute.  I have tested it with Times New Roman and it can easily be interchanged with most of the characters.  This font can be found for the time being at hangulphabet.weebly.com.  The font is in beta form at the moment
	In the future I plan to simplify the font even further.  In the works is a system that involves typing one character for manner of articulation, a second for place of articulation. Also, to simplify things even further, one could use abbreviations for place and manner of articulation.  'PL' could represent 'plosive' over the symbol BL for bilabial.  This concept has occurred to me but the Hangulphabet is less Anglo centric.   I also feel that phonological patters can be seen more clearly though symbols that show what is going on in terms of articulation.
	I am excited to see what may come of this new writing system.  In addition to easing the burden of linguists, I am positive there are a number of other needs it can be applied to.  Second language learners or literacy programs may find it useful.  This paper should serve to generate discussion and hopefully to elicit feedback that will help in further developing the system.  It will be interesting to watch both how and where the Hangulphabet is applied.  Watching it evolve will be exciting as well.

References

De Lacy, Paul ed., (2007) *The Cambridge Handbook of Phonology*. Cambridge: Cambridge Handbooks in Language and Linguistics.

Grant, B. (2002). *A Guide to Korean Characters*. New Jersey: Hollym International Corp.

*Hunmin Jeongeum Haerye*, postface of Jeong Inji, p. 27a, translation from Gari K. Ledyard, *The Korean Language Reform of 1446*, p. 258 Seoul, Korea : Singu Munhwasa, [1998]

International Phonetic Association, (1999). *Handbook of the International Phonetic Association, a Guide to the Use of the International Phonetic Alphabet*. New York, NY: Cambridge Uni. Pr.
Kopp, George A., Green, Harriet C. Potter, Ralph K., (1947). *Visible Speech*. 1st ed. New York: D. Van Nostrand Co.

Ladefoged, P., & Maddieson, I. (1996). *The Sounds of the World\'s Languages*. Wiley-Blackwell.

Ladefoged, P., & Maddieson, I. (2005) IPA Chart 2005

Sek Yen Kim-Cho, (2002). *The Korean Alphabet of 1446: Expositions Opa, the Visible Speech Sounds Translation With Annotation, Future Applicability*. 1st ed. New York: Humanity Books; Ch Op an edition.